# Fuzzy Knowledge Representation Based on Possibilistic and Necessary Bayesian Networks

ABDELKADER HENI
Preparatory Institute for
Engineering Studies
Department of Technology
Kairouan Road, 5019, Monastir
TUNISIA
abdelkader.heni@edunet.tn,

MOHAMED-NAZIH OMRI
Preparatory Institute for
Engineering Studies
Department of Technology
Kairouan Road, 5019, Monastir
TUNISIA
nazih.omri@ipeim.rnu.tn,

ADEL M ALIMI
National School of Engineers
Department of Electrical
Engineering
Soukra Road, 3000 Sfax,
TUNISIA
adel.alimi@enis.rnu.tn

*Abstract: - Within the framework proposed in this paper, we address the issue of extending the certain networks to a fuzzy certain networks in order to cope with a vagueness and limitations of existing models for decision under imprecise and uncertain knowledge. This paper proposes a framework that combines two disciplines to exploit their own advantages in uncertain and imprecise knowledge representation problems. The framework proposed is a possibilistic logic based one in which Bayesian nodes and their properties are represented by local necessity-valued knowledge base. Data in properties are interpreted as set of valuated formulas. In our contribution possibilistic Bayesian networks have a qualitative part and a quantitative part, represented by local knowledge bases. The general idea is to study how a fusion of these two formalisms would permit representing compact way to solve efficiently problems for knowledge representation. We show how to apply possibility and necessity measures to the problem of knowledge representation with large scale data. On the other hand fuzzification of crisp certainty degrees to fuzzy variables improves the quality of the network and tends to bring smoothness and robustness in the network performance. The general aim is to provide a new approach for decision under uncertainty that combines three methodologies: Bayesian networks certainty distribution and fuzzy logic*
.

## 1 Introduction

Bayesian networks have attracted much attention recently as a possible solution to complex problems related to decision support under uncertainty. These networks are systems for uncertain knowledge representation and have a big number of applications with efficient algorithms and have strong theoretical foundations [1],[2],[3],[4],[5] and [11]. They use graphs capturing causality notion between variables, and probability theory (statistic data) to express the causality power.

Although the underlying theory has been around for a long time, the possibility of building and executing realistic models has only been made possible because of recent improvements on algorithms and the availability of fast electronic computers. On the other hand, one of the main limits of Bayesian networks is necessity to provide a large number of numeric data; a constraint often difficult to satisfy when the number of random variables grows up. The goal of this paper is to develop a qualitative framework where the uncertainty is represented in possibility theory; an ordinal theory for uncertainty developed since more than ten years [6], [7], and [8]. Our framework propose to define a qualitative notion of independence (alternative to the probability theory), to propose techniques of decomposition of joined possibility distributions, and to develop some efficient algorithms for the revision of beliefs. Thus, on the first hand limitations of quantitative structure in Bayesian networks that use simple random variables have been noted by many researches. These limitations have motivated a variety of recent research in hierarchical and composable Bayesian models.

On the other hand, another limitation of the use of probabilistic Bayesian networks in expert systems is difficulty of obtaining realistic probabilities. So to solve these problems we use a new modified possibilistic Bayesian method. Our new modified possibilistic Bayesian networks simultaneously make use of both possibilistic measures: necessity measure and possibility measure.

Our work extends and refines these proposed frameworks in a number of crucial ways. The language defined in [12] [13] and in [15] has been modified to enhance usability and to support a more powerful system. We are trying in this paper to describe a language that provides the important capability of uncertainty modeling. We have also combined different element from works cited above to describe our possibilistic networks based on local necessity-valued knowledge bases.

In this paper we consider a type of possibilistic network that is based on the context model interpretation of a degree of possibility and focused on imprecision [14]. The first section presents an overview of standard possibilistic networks and their extensions. The following section describes our contribution with the use of necessity-valued knowledge bases as quantitative representation for uncertainty in nodes. And eventually we will talk about the transformations between average fuzzyBayesian networks and average knowledge bases.

## 2 Necessity-possibility measures and possibilistic networks

In order to be able to discuss our framework for possibilistic networks we shall in this section give a few preliminary definitions and notational conventions. At the same time, we present a brief outline of few important notations and ideas in possibility theory and possibilistic networks relevant to the subject of this paper.

### 2.1 Possibilistic logic
Let L be a finite propositional language. $p$; $q$; $r$; . . . denote propositional formulae.
⊤ and ⊥, respectively, denote tautologies and contradictions. ⊢ denotes the classical syntactic inference relation. $\Omega$ is the set of classical interpretations $\omega$ of L, and $[p]$ is the set of classical models of $p$ (i.e, interpretations where $p$ is true $\{\omega \mid \omega \vDash p\}$) [13].

#### 2.1.1 Possibility-necessity distributions and possibility-necessity measures
The basic element of possibility theory is the possibility distribution $\prod$ which is a mapping from $\Omega$ to the interval [0 1]. The degree $\pi(\omega)$ represents the compatibility of $\omega$ with the available information (or beliefs) about the real world. By convention, $\pi(\omega) = 0$ means that the interpretation $\omega$ is impossible, and $\pi(\omega) = 1$ means that nothing prevents $\omega$ from being the real world [13].

Given a possibility distribution $\pi$, two different ways of rank ordering formulae of the language are defined from this possibility distribution. This is obtained using two mappings grading, respectively, the possibility and the certainty of a formula p:

- The possibility (or consistency) degree:

$$\prod(p) = max\ (\pi(\omega) : \omega \in [p]) \qquad (1)$$

Which evaluates the extent to which p is consistent with the available beliefs expressed by $p$ [16]. It satisfies:

$$\forall p,\ \forall q \qquad \prod(p \vee q) = max\ (\prod(p), \prod(q)) \qquad (2)$$

- The necessity (or certainty, entailment) degree

$$N(P) = 1 - \prod(\neg p) \qquad (3)$$

Which evaluates the extent to which p is entailed by the available beliefs. We have [17]:
$$\forall p,\ \forall q \quad N(p \wedge q) = max\ (N(p), N(q)) \qquad (4)$$

To note here that in our case, we consider that both necessity degree and possibility degrees for a given formulae should be given by an expert. On the other hand, when a data is required (a possibility degree or necessity degree) one should deduce it by applying equation (3).

#### 2.1.2 Fuzzy knowledge base
A fuzzy formula is a tripley $(\varphi, \alpha, \beta)$ where $\varphi$ is a classical first-order closed formula and $(\alpha, \beta) \in [0,1]$ are a positive numbers. $(\varphi, \alpha, \beta)$ expresses that $\varphi$ is possible at least to the degree $\alpha$, and certain at least to the degree $\beta$ i.e. $\prod(\varphi) \geq \alpha$ and $\beta\ N(\varphi) \geq \beta$, where $\prod$ and $N$ are respectively a possibility and necessity measures modelling our possibly incomplete state of knowledge. The right part of a possibilistic formula, i.e. $\alpha$ and $\beta$, are respectively called the possibility and necessity *weights* of the formula.

A fuzzy knowledge base $\Sigma$ is defined as the set of weighted formulae [18]. More formally $\Sigma = \{(\varphi_t, \alpha_i, \beta_i), i = 1....m\}$ where $\varphi_t$ is a propositional formula $\alpha_i$ is the higher bound of possibility and $\beta_I$ is the lower bound of necessity accorded to this formula (certainty degree).

## 3 Fuzzy Bayesian networks
A standard possibilistic network is a decomposition of a multivariate possibility distribution according to:

$$\pi(A_1,....,A_n) = min_{i=1..n} \pi(A_i \mid parents(A_i)) \qquad (5)$$

where *parents*$(A_i)$ is the set of parents of variable $A_i$, which is made as small as possible by exploiting conditional independencies of the type indicated above [9] and [10]. Such a network is usually represented as a directed graph in which there is an edge from each of the parents to the conditioned variable.

In our work an average fuzzy Bayesian networks is considered as a graphical representation of uncertain information. It offers an alternative to probabilistic causal network when numerical data are not available.

Let $V= \{A_1, A_2,..A_n\}$ be a set of variables (i.e attributes or proprieties). The set of interpretations is the Cartesian product of all domains of attributes in *V*. When each attribute is binary, domains are denoted by $D_i=\{a_i, \neg a_i\}$.

An *average fuzzy graph* denoted by $\Pi G^A$ is an acyclic graph where nodes represents attributes i.e. a patient temperature and edges represent causal links between them. Uncertainty is represented by possibilities distribution, certainties distribution and conditional possibilities and necessities for each attribute explaining the link force between them.
The conditional possibilities and necessities distributions are associated to the graph as follow:

For each root attribute $A_i$, we specify prior possibility distribution $\Pi(a_i), \Pi(\neg a_i)$ and the prior normalization) and the prior necessity distribution $N(a_i), N(\neg a_i)$ with the constraint that :

$$\begin{cases} N(a_i) = 1 \vdash N(\neg a_i) = 0 \\ N(\neg a_i) = 1 \vdash N(a_i) = 0 \end{cases} \quad (6)$$

- For other attributes $A_j$, we specify the conditional possibilities distribution $\Pi(a_j|u_j), \Pi(\neg a_j|u_j)$ with $max(\Pi(a_i|u_j), \Pi(\neg a_i| u_j)) =1$ where $u_j$ is an instance of $a_j$ parents and the conditional necessity distribution $N(a_i), N(\neg a_i)$ with the constraint that :

$$\begin{cases} N(a_i|u_j) = 1 \vdash N(\neg a_i)|u_j = 0 \\ N(\neg a_i)|u_j = 1 \vdash N(a_i)|u_j = 0 \end{cases} \quad (7)$$

*Example*: the next figure gives an example of possibilistic Bayesian networks with four nodes and their conditional possibilities.

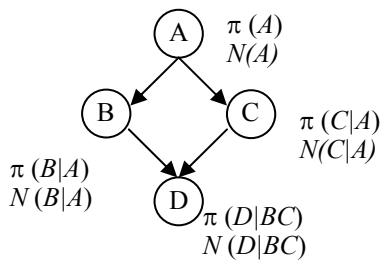

Fig. 1: example of a fuzzy Bayesian network

The *joint average distribution* is obtained then by applying the chain rule:

$A(A_1,..,A_n) = min( \pi (A_i|U(A_i)) * min (N(A_i|U(A_i))$  (8)

Where:

- $A(A_1,..,A_n)$ is The joint average distribution.
- $min ( \pi(A_i | U(A_i))$ is the lower bound of the possibilities degrees associated to $(A_i|U(A_i))$.
- $min (N(A_i|U(A_i))$ is the lower bound of the necessities degrees associated to $(A_i|U(A_i))$

*Example*: let the prior possibilities-necessities and the conditional possibilities-necessities be as described in table 1:

|     | π   | N   | A    |
|-----|-----|-----|------|
| a   | 1   | 0.6 | 0.6  |
| ¬a  | 0.5 | 0.1 | 0.05 |

|     | π | N    | π    | N   |
|-----|---|------|------|-----|
| B\|A| a | a    | ¬a   | ¬a  |
| b   | 1 | 0.5  | 0.75 | 0.2 |
| ¬b  | 0.5 | 0.25 | 0.3 | 0   |

|     | π   | N   | π   | N   |
|-----|-----|-----|-----|-----|
| C\|A| a   | a   | ¬a  | ¬a  |
| C   | 1   | 0.3 | 0.7 | 0.2 |
| ¬c  | 0.6 | 0.1 | 0.4 | 0.1 |

|      | π  | N   | π   | N   | π    | N    |
|------|----|-----|-----|-----|------|------|
| D\|BC| Bc | bc  | b¬c | b¬c | else | Else |
| d    | 1  | 0.2 | 0.5 | 0.1 | 1    | 0.3  |
| ¬d   | 0.5| 0.4 | 0.3 | 0.1 | 0.7  | 0.2  |

Table 1: possibilities-necessities distribution

By the use of the chain rule defined by equation (8) we obtain the average distribution associated with the average fuzzy Bayesian network cited above as described in table.

| A | B  | C  | D  | minΠ | minN | 𝒜    |
|---|----|----|----|------|------|------|
| a | b  | c  | d  | 1    | 0.2  | 0.2  |
| a | b  | c  | ¬d | 0.5  | 0.3  | 0.15 |
| a | b  | ¬c | d  | 0.5  | 0.1  | 0.05 |
| a | b  | ¬c | ¬d | 0.3  | 0.3  | 0.09 |
| a | ¬b | c  | d  | 0.5  | 0.2  | 0.1  |
| a | ¬b | c  | ¬d | 0.5  | 0.1  | 0.05 |
| a | ¬b | ¬c | d  | 0.5  | 0.1  | 0.05 |
| a | ¬b | ¬c | ¬d | 0.5 | 0.1 | 0.05 |
| ¬a | b | c | d | 0.5 | 0.1 | 0.05 |
| ¬a | b | c | ¬d | 0.5 | 0.1 | 0.05 |
| ¬a | b | ¬c | d | 0.4 | 0.1 | 0.04 |
| ¬a | b | ¬c | ¬d | 0.3 | 0.1 | 0.03 |
| ¬a | ¬b | c | d | 0.3 | 0 | 0 |
| ¬a | ¬b | c | ¬d | 0.3 | 0 | 0 |
| ¬a | ¬b | ¬c | d | 0.3 | 0 | 0 |
| ¬a | ¬b | ¬c | ¬d | 0.3 | 0 | 0 |

Table 2: joint average possibility-necessity distribution

## 4 Average possibilistic and necessary valued knowledge base

We would like to represent a class of possibilistic Bayesian networks using a local average fuzzyvalued knowledge base consisting of a collection of possibilistic logic sentences (formulae) in such a way that a network generated on the basis of the information contained in the knowledge base is isomorphic to a set of ground instances of the formulae. As the formal representation of the knowledge base, we use a set of possibilistic formulae. We represent random variables with necessities and possibilities weights and restrict ourselves to using only the average of these two measures.

Formally an average necessity-possibility valued knowledge base is defined as the set :

$$\Sigma = \{(\varphi_t, \alpha_i, \beta_i), i = 1 \ldots m\} \quad (9)$$

Where $\varphi_t$ denotes a classical propositional formula, $\alpha_i$ and $\beta_i$ denote respectively the lower bound of certainty (i.e necessity) and the lower bound of possibility.

We can represent the information contained in each node of a Bayesian network, as well as the quantitative information contained in the link matrices, if we can represent all the direct parent/child relations. We express the relation between each random variable and its parents over a class of networks with a collection of quantified formulae. The collection of formulae represents the relation between the random variable and its parents for any ground instantiation of the quantified variables. The network fragment consisting of a random variable and its parents with a set of formulae of the form $(\varphi, \alpha, \beta)$.

We give next some definitions inspired from [12] and [13].

*Definition 1*:
Two average knowledge bases $\Sigma^A_1$ and and $\Sigma^A_2$ are said to be equivalent if their associated possibility distributions (respectively necessity distributions) are equal, namely:

$$\begin{cases} \forall \omega \in \Omega, \; \pi\Sigma^A_1(\omega) = \pi\Sigma^A_2(\omega) \\ \text{and} \\ \forall \omega \in \Omega, \; N\Sigma^A_1(\omega) = N\Sigma^A_2(\omega) \end{cases} \quad (10)$$

*Definition 2*:
Let $(\varphi, \alpha, \beta)$ a formula in $\Sigma^A$ Then $(\varphi, \alpha, \beta)$ is said to be subsumed by $\Sigma^A$ if $\Sigma^A$ and $\Sigma^A \setminus \{(\varphi, \alpha, \beta)\}$ are equivalent knowledge bases.

This is means that each redundant formula should be removed from the average valued knowledge base since it can be deduced from the rest of formulae.

## 5 From fuzzy Bayesian network to fuzzy knowledge base

In this section, we describe the process that permit to deduce an average valued knowledge base from an average network.

Let $\Pi G^A$ be an average and necessary Bayesian network consisting of a set of labeled variables $V = \{A_1, A_2, \ldots A_n\}$. Now let $A$ be a binary variable and let $(a \; \neg a)$ be its instances.
Given the two measures $\pi(a_i|u_i)$ and $N(a_i|u_i)$ witch represent respectively the local possibility degree and the local necessity degree associated with the variable $A$ where $u_i \in U_A$ is an instance of parents($a_i$). the local average knowledge base associated with $A$ should be defined using the next equation :

$$\Sigma^A_A = \{(\neg a_i \vee u_i, \alpha_t, \beta_i), \alpha_t = 1 - \pi(a_i|u_i) \neq 0 \text{ and } \beta_i = 1 - N(a_i|u_i) \neq 0 \} \quad (11)$$

To note here that in [15] the authors prove the possibility to recover conditional possibilities from $\Sigma_A$ where $\Sigma_A$ is a possibilistic knowledge base.

Based o the results obtained in [15], we can check in our case that it is possible to recover both conditional necessities from $\Sigma^A_A$ according to equations (12) and (13).

$$\Pi\Sigma^A(\omega) = \begin{cases} 1 & \text{if } \forall \; (\varphi_i, \alpha_i) \in \Sigma \; \omega \models \varphi_i \\ 1 - \max\{\alpha_i : \omega \not\models \varphi_i\} & \text{otherwise} \end{cases} \quad (12)$$

*and*

$$N\Sigma^A(\omega) = \begin{cases} 1 & \text{if } \forall\ (\varphi_i, \alpha_i) \in \Sigma\ \omega \vDash \varphi_i \\ 0 & \text{otherwise} \end{cases} \quad (13)$$

*Example*: by applying equation (11), we get the average knowledge base associated to the average fuzzy Bayesian network described in section 3.

$\Sigma^A_A = \{(a, 0.5, 0.9\ )\}\quad = \{(a, 0.45\ )\}$
$\Sigma^A_B = \{(b \vee a, 0.7), (b \vee \neg a, 0.5, 0.75)(\neg b \vee a, 0.25, 0.8)\}$
$\quad = \{(b \vee a, 0.7), (b \vee \neg a, 0.375)\ (\neg b \vee a, 0.2)\}$

$\Sigma^A_C = \{(c \vee a, 0.6, 0.9), (c \vee \neg a, 0.4, 0.9)\ (\neg c \vee a, 0.3, 0.8)\}$
$\quad = \{(c \vee a, 0.54), (c \vee \neg a, 0.36\ 0.9)\ (\neg c \vee a, 0.24)\}$

$\Sigma^A_D = \{\{(d \vee b \vee c, 0.3, 0.8), (d \vee b \vee \neg c, 0.3, 0.8), (d \vee \neg b \vee c, 0.7, 0.9), (d \vee \neg b \vee \neg c, 0.5, 0.6\ ), (\neg d \vee \neg b \vee c, 0.5, 0.9\ )\}\}$
$\quad = \{\{(d \vee b \vee c, 0.24), (d \vee b \vee \neg c, 0.24), (d \vee \neg b \vee c, 0.63), (d \vee \neg b \vee \neg c, 0.3\ ), (\neg d \vee \neg b \vee c, 0.45\ )\}\}$

*Remark*: for each knowledge base the first equality represents the initial knowledge base weighted by possibilities and necessities when the other represents the average based knowledge base (namely average necessity-possibility valued knowledge base).

Next section shows the other face of transformation between average valued knowledge base and average fuzzy Bayesian network.

## 6 From Average valued knowledge base to average fuzzy Bayesian network

In [15] the authors describe a process permitting to deduce a possibilistic network from a possibilistic knowledge base. In this section we follow the same way to transform our average necessity-valued knowledge bases into an average fuzzy Bayesian network.

To note here that the average possibilistic and necessary Bayesian network deduced from an average necessity-valued knowledge bases will have the same graphical structure as the starting network

The conditional average distributions are simply the ones associated with the average knowledge bases. More precisely, let $A_i$ be variable and $u_i$ be an element of parents($A_i$). Let $\Sigma^A_{Ai}$ be the local average knowledge base associated with the node $A_i$. Then, the conditional average degree $A(a_i|u_i)$ is defined by $\pi(a_i|u_i) = \pi(a_i \wedge u_i)$
$= \pi \Sigma^A_{Ai}(a_i \wedge u_i)$ and $\Sigma^A_{Ai}(a_i \wedge u_i)$ is defined using equation (12) and equation (13).

Respectively the conditional necessity degree $N(a_i|u_i)$ is defined by $N(a_i|u_i) = N(a_i \wedge u_i) = N\Sigma^A_{Ai}(a_i \wedge u_i)$.
*Example:*

From the average knowledge base associated to the node $A$ and by the use of equations 11 and 12
$\Sigma^A_A = \{(a, 0.5, 0.9\ )\}$
$\quad = \{(a, 0.45\ )\}$
We can deduce the conditional average table for node $A$ by the use of equations 11 and 12

|     | π   | N   | A    |
|-----|-----|-----|------|
| a   | 1   | 0.6 | 0.6  |
| ¬a  | 0.5 | 0.1 | 0.05 |

Same to rest of nodes we can deduce the rest of conditional averages associated to other nodes and so we can recover the average distribution presented in table 2.

## 7 Fuzzy Bayesian networks based on fuzzy necessity distribution

Logical formulae with a weight strictly greater than a given levels (lower bounds of necessity degrees) are immune to inconsistency and can be safely used in deductive reasoning [19]. However in order to perform reasoning for both imprecise and uncertain information, two important issues should be addressed. First, any improvement of the possibility level for a piece of information can only be achieved at the expense of the specificity of the information; second the accorded levels to the causality explained in terms of rules (case of fuzzy logic) and conditional dependencies (case of Bayesian networks) are somewhat expensive due to the fact that these confidence level is somewhat critical.

We propose so to combine these three approaches (Bayesian networks certainty distribution and fuzzy logic) to develop a method for uncertain and imprecise knowledge representation that may improve decision based systems.

Our fuzzy beliefs are to emulate a certain Bayesian necessity measure. For simplicity each variable here has two states: the presence or absence of an entity. The belief that A is present takes the form of a fuzzy truth fA. The extent to witch the belief of variable state influences the state beliefs of parent or child is modelled by a fuzzy set membership function: one for each influence direction.

*Example:*
Let our certain network be as described in figure representing a Bayesian network in metastatic cancer.

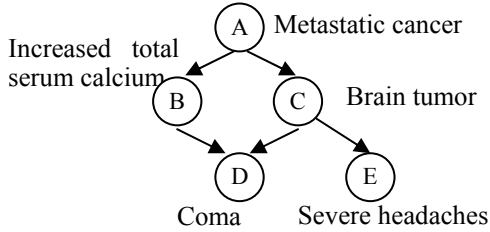

*Fig 2. A Bayesian network for metastatic cancer[20]*

Fig. 2 shows a Bayesian network representing the above cause and effect relationships. Table 3 lists the causal influences in terms of fuzzy certainty distributions. Each variable is characterized by an unknown necessity degree given the state of its parents. For instance: $C \in [0, 1]$ represents the dichotomy between having a brain tumor and not having one, $c$ denotes the assertion $C = 1$ or "Brain tumor is present", and $\neg c$ is the negation of $c$, namely, $C = 0$. The root node, $A$, which has no parent, is characterized by its prior fuzzy certainty distribution.

*Example*
Le the conditional fuzzy necessities associated to the graph presented in figure 2 be as described in table 3. For reason of simplicity we kept here four nodes only as in the graph presented in figure 1.

| A | ¬a |
|---|---|
| $[\beta_{A11}, \beta_{A12}]$ | $[\beta_{A21}, \beta_{A22}]$ |

| B\|A | A | ¬a |
|---|---|---|
| b | $[\beta_{B|A11}, \beta_{B|A12}]$ | $[\beta_{B|A21}, \beta_{B|A12}]$ |
| ¬b | $[\beta_{B|A31}, \beta_{B|A32}]$ | $[\beta_{B|A41}, \beta_{B|A42}]$ |

| C\|A | a | ¬a |
|---|---|---|
| c | $[\beta_{C|A11}, \beta_{C|A12}]$ | $[\beta_{C|A21}, \beta_{C|A12}]$ |
| ¬c | $[\beta_{C|A31}, \beta_{C|A32}]$ | $[\beta_{C|A41}, \beta_{C|A42}]$ |

| D\|BC | bc | b¬c | Else |
|---|---|---|---|
| d | $[\beta_{D|BC11}, \beta_{D|BC12}]$ | $[\beta_{D|BC21}, \beta_{D|BC22}]$ | $[\beta_{D|BC31}, \beta_{D|BC32}]$ |
| ¬d | $[\beta_{D|BC41}, \beta_{D|BC42}]$ | $[\beta_{D|BC51}, \beta_{D|BC52}]$ | $[\beta_{D|BC61}, \beta_{D|BC62}]$ |

Table 3:  fuzzy necessity distribution

For instance N(d| b,¬c) cannot be 0.1 as described in table 1 but rather is a fuzzy number say $\chi_1 \in [\beta_{D|BC1}, \beta_{D|BC2}]$ where $\chi_1 = \aleph(d| b,¬c)$ is the fuzzy necessity associated with the fuzzy formula (d| b,¬c) and is associated with a membership function $\mu(\chi_1)$ supposed to be a triangular function (respectively $\mu$ can be trapezoid or other kind of functions). $\mu$ is represented as follow (figure 3):

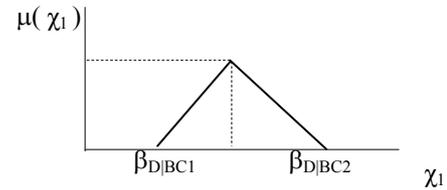

*Fig. 3  a membership function*

Then we can deduce the next possible representation of $\mu(\chi_1)$ as:

$$\mu(\chi_1) = k_1 \times (\chi_1 - \beta_{D|BC1}) - k_2 \times (|\chi_1 - \beta_{D|BC2}| + \chi_1 - \alpha)$$

Where:
- $\alpha$, $k_1$ and $k_2$ are two defined constants..
- $|*|$ is the absolute value of term $*$

The above expression and figure mean that the interval of $\chi_1$ is $[\beta_{D|BC1}, \beta_{D|BC2}]$. If $\chi_1 = \alpha$ then $\mu(\chi_1)=1$, implying that the fuzzy necessity $\chi_1 = \alpha$ is the most possible situation. If $\chi_1 \geq \beta_{D|BC2}$ or $\chi_1 \leq \beta_{D|BC1}$ then $\mu(\chi_1) = 0$, the possible manifestation of $\chi_1$.

## 8. Transformation between FBN and fuzzy Knowledge bases

Analogously, when the given necessities degree are fuzzy numbers as we described in section 5, the necessity distribution N(X) associated to a node X is considered as a fuzzy distribution defined by a membership function

$$\mu : [\beta_1, \beta_2] \longrightarrow [0\ 1] \quad (14)$$
$$\chi \longmapsto \mu(\chi)$$

Example: consider the graph of figure 2. For simplicity each variable here has two states: the presence or absence of an entity and we will define the same membership function to a as to ¬a.

| a | ¬a |
|---|---|
| $[\beta_{A11}, \beta_{A12}]$ | $[\beta_{A21}, \beta_{A22}]$ |
| $\mu_1(\chi)$ | $\mu_1(\chi)$ |

| B\|A | a | ¬a |
|---|---|---|
| b | [$\beta_{B|A11}$, $\beta_{B|A12}$] | [$\beta_{B|A21}$, $\beta_{B|A12}$] |
|   | $\mu_2(\chi)$ | $\mu_3(\chi)$ |
| ¬b | [$\beta_{B|A31}$, $\beta_{B|A32}$] | [$\beta_{B|A41}$, $\beta_{B|A42}$] |
|   | $\mu_2(\chi)$ | $\mu_3(\chi)$ |

| C\|A | a | ¬a |
|---|---|---|
| c | [$\beta_{C|A11}$, $\beta_{C|A12}$] | [$\beta_{C|A21}$, $\beta_{C|A12}$] |
|   | $\mu_4(\chi)$ | $\mu_5(\chi)$ |
| ¬c | [$\beta_{C|A31}$, $\beta_{C|A32}$] | [$\beta_{C|A41}$, $\beta_{C|A42}$] |
|   | $\mu_4(\chi)$ | $\mu_5(\chi)$ |

| D\|BC | bc | b¬c | Else |
|---|---|---|---|
| d | [$\beta_{D|BC11}$, $\beta_{D|BC12}$] | [$\beta_{D|BC21}$, $\beta_{D|BC22}$] | [$\beta_{D|BC31}$, $\beta_{D|BC32}$] |
|   | $\mu_6(\chi)$ | $\mu_7(\chi)$ | $\mu_8(\chi)$ |
| ¬d | [$\beta_{D|BC41}$, $\beta_{D|BC42}$] | [$\beta_{D|BC51}$, $\beta_{D|BC52}$] | [$\beta_{D|BC61}$, $\beta_{D|BC62}$] |
|   | $\mu_1(\chi)$ | $\mu_7(\chi)$ | $\mu_8(\chi)$ |

Table 4: fuzzy necessity distribution with membership functions

Let the different membership be as follow:

$\mu_i(\chi) = k_{i1} \times (\chi - \beta_{ij1}) - k_{i2} \times (|\chi - \beta_{ij2}| + \chi - \alpha_i)$

Where:
- $\mu_i(\chi)$ is the membership function associated to the fuzzy variable $\chi$, supposed to be triangular.

- $k_{i1}$ and $k_{i2}$ are the used constant in each membership function supposed to be triangular.

- $\beta_{ij1}$ and $\beta_{ij2}$ are the two min and the max boundary of a necessity degree.

Finally by maximization of each membership function, we can deduce an optimal value for the certainty degree associated to each fuzzy variable (i.e. proposition). Namely:

$\aleph(\chi) = \mu(\chi) = 1$

Then it will be easy to deduce the value of $\chi$ as follow:

$$\chi = \frac{\lambda + k_{i1} \times \beta_{ij1} + k_{i2} \times \beta_{ij2} + 1}{k_{i1}} \qquad (15)$$

By replacing $\lambda$ by 1 (the maximization of $\mu(\chi)$), the value of $\chi$ will be:

$$\chi = \frac{\lambda + k_{i1} \times \beta_{ij1} + k_{i2} \times \beta_{ij2} + 1}{k_{i1}} \qquad (16)$$

Analogously, the definition of the fuzzy joint necessity distribution is obtained by applying the fuzzy chain rule:

$\aleph(A_1,...,A_n) = \min(\chi_i)$, $\chi_i = \aleph(A_i|U(A_i))$

From a semantic point of view, a certain knowledge base $\Sigma = \{(\varphi_i, \alpha_i), i = 1....m\}$ where each $\alpha_i$ a crisp necessity value, is understood as the necessity distribution $N\Sigma$ representing the fuzzy sets of models of $\Sigma$:

$N\Sigma(\omega) = \min \max (\mu_{[Pi]}(\omega), 1-\alpha)$ where $[P_i]$ denotes the set of models of $P_i$, so that :

$$\mu_{[Pi]}(\omega) = \begin{cases} \mu_{[Pi]} = \alpha & \text{if } \omega \in P_i \\ 0 & \text{otherwise} \end{cases} \qquad (17)$$

From (21) we can clearly deduce clearly that $N\Sigma(\omega)$ is naturally a fuzzy distribution applied to a crisp set of values and $\mu_{[Pi]}$ is the crisp membership function.

## 9 Conclusion

This paper has presented a definition of fuzzy Bayesian networks and how to use them to deduce average knowledge bases and vice versa. Uncertainty in nodes in our models is represented by local knowledge bases.

The key benefits of this representation to the practitioner are that both knowledge declaration and possibilistic inference are modular. Individual knowledge bases should be separately compilable and query complete. Also this representation specifies an organized structure for elicitation of the graph structure. We only defined the transformation process for knowledge bases.

Certain Bayesian networks with fuzzy knowledge bases approach in a natural way gives us the subsethood of the evidence for each logical formula. Although the methodology proposed in this paper, is aimed and illustrated by some typical examples, the developed techniques require experimental results.

A future work is to extend this representation by definition of efficient algorithms for locally inferences.